\documentclass[conference]{IEEEtran}
\IEEEoverridecommandlockouts
\usepackage[T1]{fontenc}
\usepackage[utf8]{inputenc}
\usepackage[english]{babel}
\usepackage{caption}
\usepackage{subcaption}
\usepackage{pgfplots}
\usepackage{svg}
\pgfplotsset{compat=1.16}

\usepackage{cite}
\usepackage{amsmath,amssymb,amsfonts}
\usepackage{algorithmic}
\usepackage{booktabs}
\usepackage{textcomp}
\usepackage{xcolor}
\usepackage[colorinlistoftodos,prependcaption,textsize=tiny]{todonotes} 
\usepackage{xargs}
\usepackage{graphicx}
\graphicspath{{figures/}}
\usepackage{hyperref}
\hypersetup{
    colorlinks=true,
    linkcolor=black,
    citecolor=black,
    filecolor=black,
    urlcolor=black,
}
\usepackage{cleveref}

\usepackage{xfrac}
\usepackage{siunitx} 
\newcommandx{\unsure}[2][1=]{\todo[linecolor=red,backgroundcolor=red!25,bordercolor=red,#1]{#2}}
\newcommandx{\change}[2][1=]{\todo[linecolor=blue,backgroundcolor=blue!25,bordercolor=blue,#1]{#2}}
\newcommandx{\info}[2][1=]{\todo[linecolor=OliveGreen,backgroundcolor=OliveGreen!25,bordercolor=OliveGreen,#1]{#2}}
\newcommandx{\improvement}[2][1=]{\todo[linecolor=Plum,backgroundcolor=Plum!25,bordercolor=Plum,#1]{#2}}
\newcommandx{\thiswillnotshow}[2][1=]{\todo[disable,#1]{#2}}

\newcommand*{\diff}[1]{\mathop{}\!\mathrm{d}{#1}\!\mathop{}}
\newcommand{\x}{\mathbf{x}}
\renewcommand{\u}{\mathbf{u}}
\renewcommand{\L}{\text{L}}
\newcommand{\step}{\tau}
\newcommand{\pita}{\text{PITA}}

\def\BibTeX{{\rm B\kern-.05em{\sc i\kern-.025em b}\kern-.08em
    T\kern-.1667em\lower.7ex\hbox{E}\kern-.125emX}}

\begin{document}

\title{PITA: Physics-Informed Trajectory Autoencoder\\
{}
\thanks{*equal contribution, names ordered alphabetically}
\thanks{$^{1}$Karlsruhe Institute for Technology {\{firstname.lastname\}@kit.edu}}%
\thanks{$^{2}$Research Center for Information Technology}%
}

\author{Johannes Fischer$^{*,1}$,
Kevin Rösch  $^{*,1,2}$,
Martin Lauer$^{1}$,
Christoph Stiller$^{1}$
}

\maketitle

\begin{abstract}

    Validating robotic systems in safety-critical applications requires testing
    in many scenarios including rare edge cases that are unlikely to occur,
    requiring to complement real-world testing with testing in simulation.
    Generative models can be used to augment real-world datasets with generated data
    to produce edge case scenarios
    by sampling in a learned latent space.
    Autoencoders can learn said latent representation for a specific domain
    by learning to reconstruct the input data from a lower-dimensional intermediate representation.
    However, the resulting trajectories are not necessarily physically plausible,
    but instead typically contain noise that is not present in the input trajectory.
    To resolve this issue,
    we propose the novel Physics-Informed Trajectory Autoencoder (PITA) architecture,
    which incorporates a physical dynamics model into the loss function of the autoencoder.
    This results in smooth trajectories that not only
    reconstruct the input trajectory
    but also adhere to the physical model.
    We evaluate PITA on a real-world dataset of vehicle trajectories
    and compare its performance to a normal autoencoder
    and a state-of-the-art action-space autoencoder.

\end{abstract}

\begin{IEEEkeywords}
Autoencoder, Trajectory, Physics-informed, Loss, Generative Model, Data Augmentation
\end{IEEEkeywords}
\section{Introduction}

The introduction of robotic systems into domains that were previously operated by humans
enables a wide range of new applications,
such as autonomous driving, robotic surgery, and automated manufacturing.
However, these systems need to be able to operate safely and reliably in a wide range of scenarios,
and their safety has to be proven before they can be deployed.
This requires extensive testing and validation in a wide range of scenarios,
in particular, including rare or dangerous edge cases that are difficult to encounter in real-world testing.

To address this challenge, generative models have been proposed
to generate data for testing of robotic systems.
These systems can be used to construct edge cases by augmenting a real-world dataset with generated data.
Earlier generative models did not use an encoder to generate data,
for instance, DALL-E generates images from text descriptions without using an encoder
\cite{rameshZeroShotTexttoImageGeneration2021}.
More recently, generative models have been proposed that use an encoder in their architecture.
For example, the Stable Diffusion model uses an encoder extracted from an autoencoder trained in an unsupervised manner
to generate high-resolution images from a low-dimensional latent space conditioned on context information
\cite{rombachHighResolutionImageSynthesis2022}.
Stable diffusion has been empirically shown to outperform
previous generative models in terms of synthesized image quality.

Our goal is to develop an encoder that captures the physical properties of trajectories
and hence is suitable for trajectory generation.
To this end, we investigate trajectory autoencoders
which encode trajectories in a lower-dimensional latent space.
The autoencoder is trained to reconstruct the input trajectory from the latent space,
and can be used to generate new trajectories by sampling from the latent space.
While being out of the scope of this work,
the trained encoder can be used to construct a more involved generative trajectory model
based on the learned latent space representation.

\begin{figure}
	\centering
	\includesvg[width=\linewidth]{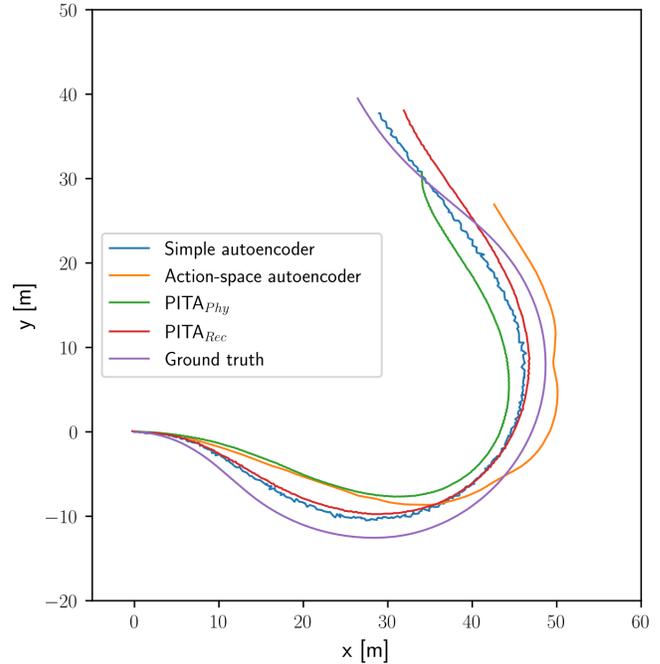}
	\caption{Comparison of different trajectory reconstruction methods resulting in more or less smooth trajectories.}
	\label{fig:reconstruction_example}
\end{figure}

The problem with traditional trajectory generation is that the resulting trajectories are not necessarily
physically plausible.
The loss function is usually defined as the mean squared error between the input and the output trajectories.
This results in a rather good reconstruction of the input trajectory,
but the output trajectory still contains jitter and noise that is not present in the input trajectory
and make it non-smooth,
as the example in \cref{fig:reconstruction_example} illustrates.
This is particularly problematic if the predicted points have a high temporal resolution,
as is required for generating edge cases:
The motion between two adjacent points will not be physically plausible due to the noise-induced jitter.
The training process does not capture the underlying physics of the system,
but rather learns to reproduce the points almost independently of each other.

To address this issue, we propose to incorporate a physical model into the loss function of the autoencoder.
The physical model captures the system's motion dynamics
to ensure that the predicted trajectories are smooth and physically plausible.
This gives rise to the physical loss, measuring how much the predicted trajectory
violates the physical model.
The loss function is defined as a combination of the reconstruction loss and the physical loss.
Besides the loss function, the whole autoencoder architecture remains unchanged.
This implies that our proposed loss can be added to any existing network architecture to improve the physical plausibility of the generated trajectories.
The contributions of our work are threefold:
\begin{itemize}
	\item We propose the Physics-Informed Trajectory Autoencoder (PITA),
	which uses a novel loss function that incorporates a physical model of the robotic system.
	\item We evaluate PITA on a real-world dataset of vehicle trajectories
	and propose metrics for evaluating the smoothness of a trajectory.
	\item We compare our approach to a traditional autoencoder
	and a state-of-the-art action-space based autoencoder \cite{janjosSelfSupervisedActionSpacePrediction2021}.
\end{itemize}

\section{Related Work}
\subsection*{Action-Space Prediction}

Trajectory prediction is a well-established research area in robotics.
Action-space prediction is an approach that allows to consider a dynamics model of the system
\cite{janjosSelfSupervisedActionSpacePrediction2021, janjosSANSceneAnchor2022}.
Those approaches use machine learning models to predict the future control inputs of the system.
These control inputs are then used to integrate the dynamics model to predict the future trajectory.
The model is trained to minimize the mean squared error between
the resulting integrated trajectory and the ground truth trajectory.
As a result, the prediction is implicitly guaranteed to satisfy the dynamics model.

\subsection*{Physics-Informed Neural Networks}

An alternative to incorporate physical models into the learning process is given by
Physics-Informed Neural Networks (PINNs) \cite{raissiPhysicsinformedNeuralNetworks2019}.
These models are trained to output the quantities of interest directly,
while using a physical relationship as a regularization.
This way, the neural network learn to output predictions that satisfy the physical model.

Physics-informed machine learning can benefit training with limited data,
since the physical model provides additional information in between data points.
Similarly, it also aids generalization to unseen scenarios,
since the physical relation can be extrapolated to new situations.

Previous work has used physics-informed machine learning to incorporate driver models
into imitation learning
\cite{moPhysicsInformedDeepLearning2021,naingPhysicsguidedGraphConvolutional2023}
and reinforcement learning
\cite{fischerPhysicsinformedReinforcementLearning2023}
for automated driving.
The policies were trained to perform well, but also to be similar to the driver model guidance.
This resulted in better performance on unseen data,
thus alleviating the problem of covariate shift.
In contrast to our work, the physical models these works used are not actual models derived from vehicle kinematics,
but rather proxy models for car-following behavior.

Another work uses the kinematic bicycle model to
\cite{maheshwariPIAugPhysicsInformed2023}
learn vehicle dynamics for off-road terrain.
A neural network is trained to predict the future trajectory using
a regression loss for predicted points and a physics loss to make the predictions
adhere to the kinematic bicycle model.
They use the distance between predictions from the neural network and predictions from the bicycle model as physical loss.
This is in contrast to our work,
where the physical loss is composed of the error in the differential equation
that described the kinematic bicycle model,
and hence also penalizes non-smooth predictions.


\section{Method}

\subsection{Physical Model}

To achieve smooth trajectory predictions,
we use the three degree-of-freedom kinematic bicycle model (KBM)
as a physical regularization.
Since we do not consider datasets with high-speed maneuvers,
the KBM is sufficient to capture the vehicle dynamics.
Usually, the KBM is formulated with the state $\x$ consisting of
the vehicle's Cartesian position $(x, y)$,
the heading angle $\theta$ and the velocity $v$.
The control inputs $\u$ are the steering angle $\delta$ and the acceleration $a$.
The KBM relates the state and control inputs to the state derivatives
using the wheelbase.

Since we train our approach on a real-world recorded dataset of vehicle trajectories
from different vehicles, we do not have access to each vehicle's wheelbase and steering angle.
For this reason, we use the curvature $\kappa$ as an input variable
instead of the steering angle $\delta$.
This allows to formulate motion model independent of the vehicle's wheelbase as
\begin{equation}
\frac{\diff{\x}}{\diff{t}}
=\frac{\diff{}}{\diff{t}} \begin{bmatrix} x \\ y \\ \theta \\ v \end{bmatrix}
= \begin{bmatrix} v \cos(\theta) \\ v \sin(\theta) \\ v \kappa \\ a \end{bmatrix}
= f(\x, \u)
\end{equation}
using the state $\x=(x,y,\theta,v)$ and control inputs $\u=(\kappa, a)$.

\subsection{Loss definition}
\label{sec:loss_definition}
\textbf{Reconstruction loss}
For one trajectory of length $T$,
the reconstruction loss is given by the squared distance of corresponding
predicted and ground truth trajectory points.
\begin{equation}
\L_{\text{Rec}} = \sum_{t=1}^{T}
\left\| \begin{pmatrix} \hat{x}_t \\ \hat y_t \end{pmatrix} - \begin{pmatrix} {x}_t \\ y_t \end{pmatrix}\right\|^2
\end{equation}

\textbf{Physical loss}
\begin{align}
\L_{\text{Phy}}
=& w_1\left\|\frac{\diff{\hat{x}}}{\diff{t}}-\hat{v} \cos\hat{\theta}\right\|^2
+ w_2\left\|\frac{\diff{\hat{y}}}{\diff{t}}-\hat{v} \sin\hat{\theta}\right\|^2
\label{eq:physical_loss_1}\\
+& w_3\left\|\frac{\diff{\hat{\theta}}}{\diff{t}}-\hat{v}\hat{\kappa}\right\|^2
+ w_4\left\|\frac{\diff{\hat{v}}}{\diff{t}}-\hat{a}\right\|^2
\label{eq:physical_loss_2}\\
+& w_5\|\hat\kappa\|^2
+  w_6\|\hat a\|^2
\label{eq:physical_loss_3}
\end{align}

The terms in \cref{eq:physical_loss_1,eq:physical_loss_2} are the errors in the differential equation,
while the terms in \cref{eq:physical_loss_3} are regularization terms to prevent the model from using excessive accelerations and curvatures.

\textbf{Loss weighting:}
Since the terms of the physical loss have different units,
we need to scale them to have similar magnitudes.
To this end, we empirically tune the weights $w_1,\ldots,w_6$
to balance the contributions of the different terms in the physical loss.

Additionally, we also need to balance the contributions
of the reconstruction loss and the physical loss.
To this end, we introduce hyperparameters $\lambda_1$ and $\lambda_2$ for the two loss term.

In preliminary experiments,
we observed that the model is prone to collapsing to zero predictions if
$\lambda_2$ is too large, since a zero prediction results in zero physical loss.
To resolve this, we have implemented a scheduling scheme for the physical loss,
where we start with a small value and increase it exponentially over time during training,
until it reaches its maximum value.
The employed scheduling scheme is given by
\begin{equation}
	\alpha(\step) = \min\left(1, e^{m\left(\frac{\step}{\gamma\step_{\text{max}}}-1\right)}\right)
\end{equation}
where $\step$ is the current training step,
$\step_{\text{max}}$ is the maximum number of training steps,
$m$ is a scaling factor which influences the shape of the increase
and $\gamma$ is the fraction of training steps after which the parameter reaches its maximum value.

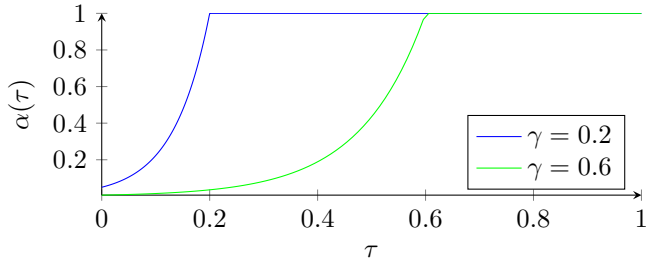
\begin{figure}[t]
\begin{tikzpicture}[trim left=-1.3cm]
\begin{axis}[
	axis lines = left,
	xlabel = $\step$,
	ylabel = {$\alpha(\step)$},
	width=\linewidth - 0.1cm,
	height=4cm,
	legend pos=south east,
]

\addplot [
	domain=0:1,
	samples=1000,
	color=blue,
]
{min(1, 1 / exp(3*(1 - x / (0.2*1))))};

\addplot [
	domain=0:1,
	samples=100,
	color=green,
]
{min(1, 1 / exp(5*(1 - x / (0.6*1))))};

\legend{$\gamma=0.2$, $\gamma=0.6$}
\end{axis}
\end{tikzpicture}
\caption[scheduling]{Physical loss weight scheduling scheme for different parameter values of $\gamma$.}
\label{fig:scheduling}
\end{figure}

\Cref{fig:scheduling} illustrates the weight scheduling scheme $\alpha$
for different parameters values with $m=5$ and $\gamma\in\{0.2, 0.6\}$.
The final trajectory loss in training step $\step$ is then given by
\begin{equation}
	\label{eq:total_loss}
\L(\step) = \lambda_1 \L_{\text{Rec}} + \alpha(\step)\lambda_2 \L_{\text{Phy}}.
\end{equation}
\section{Evaluation}

\subsection{Experimental Setup}
\label{sec:experimental_setup}

\textbf{Dataset:}
To evaluate our method, we use a real-world dataset of vehicle trajectories.
Different trajectory datasets of preprocessed drone recordings are available to the research community
\cite{krajewskiHighDDatasetDrone2018,krajewskiDatasetDroneDataset2020,zhanINTERACTIONDatasetINTERnational2019}.
To be able to augment data with generated trajectories at a high frame rate,
we use a dataset that contains trajectories with high temporal resolution.
Since we are interested in how well our model can capture the vehicle dynamics model,
highway trajectories that are mostly straight are of less interest.
For this reason, we chose the rounD dataset for our evaluation,
which contains vehicles trajectories in roundabouts \cite{krajewskiDatasetDroneDataset2020}.
More precisely, it is composed of 13509 vehicle trajectories at three different locations with a total recorded time of
over six hours, sampled with \SI{25}{\hertz},
i.e.\ at time steps of $\Delta t = \SI{0.04}{\second}$.

\textbf{Preprocessing:}
A data point for our training is made up of a trajectory of one vehicle.
Generally, it would be interesting to reconstruct the trajectories for all vehicles
in the traffic scene given the map context.
However, since our main goal is to investigate the benefits
of incorporating a physical model into the learning process,
we chose to use single vehicle trajectories as input for the autoencoder.
The trajectories are represented by the Cartesian positions of the vehicle at each time step.
We translate all trajectories to start at the origin and rotate them to have the same initial orientation.
As our autoencoder model has a fixed input dimension,
we cut the trajectories to a fixed length of $T=350$ time steps,
which corresponds to a duration of 14 seconds.
Since each point consists of x- and y-position, our model has an input dimension of $2\cdot T=700$.

\textbf{Autoencoder architectures:}
We compare three different autoencoder architectures,
which share the same architecture,
but differ only in their output representations and loss functions.
The autoencoder model follows the traditional network layout, we chose a depth of 12 layers for the encoder and decoder.
The latent space has a fixed dimension of 320. This results in a total of round about 8 million parameters.
\begin{enumerate}
	\item \textit{Simple autoencoder:}
	      The first baseline is a regular autoencoder
	      where the output has the same dimension as the input
	      and represents the reconstructed trajectory positions.
	      It is trained to minimize the mean squared error between the input trajectory positions
	      and the predicted trajectory positions.
	\item \textit{Action-space autoencoder:}
	      The second baseline is an action-space autoencoder,
	      based on the architecture proposed by Janjos et al. \cite{janjosSelfSupervisedActionSpacePrediction2021}.
	      For the action-space autoencoder, the model output has a dimension of $2\cdot T + 4$.
	      The first four entries represent the predicted initial state of the vehicle
	      $\hat\x_0=(\hat x_0,\hat y_0,\hat \theta_0,\hat v_0)$
	      while the remaining elements are the predicted control inputs
	      $\hat\u_t=(\hat\kappa_t,\hat a_t)$
	      for the dynamics model at each time step $t$.
	      Using these predicted control inputs,
	      we integrate the vehicle dynamics model to obtain the predicted state trajectory
	      using a fourth order Runge-Kutta method starting from the predicted initial state.
	      The loss function is defined as the mean squared error between the input trajectory
	      and the positions of the predicted state trajectory.
	\item \textit{Physics-informed autoencoder:}
	      Our physics-informed trajectory autoencoder (PITA)
	      has an output dimension of $6\cdot T$,
	      representing the predicted state and control input trajectory $(\x_t,\u_t)_{t=1}^T$.
	      Based on the predicted states and control inputs, we compute the
	      total loss as the weighted sum of the reconstruction loss between
	      input trajectory and predicted positions
	      and the physical loss
	      as described in \cref{sec:loss_definition}.
	      To compute the physical loss,
	      we approximate the state derivatives using central finite differences on the model outputs.
\end{enumerate}

\textbf{Training:}
All networks are trained with trajectories provided by the dataset described in \cref{sec:experimental_setup}.
The dataset is split into a training and a validation set,
where we use the same split for all training runs to ensure comparability.
In each training epoch,
we train on randomly samples mini-batches
using the mean squared error over a training batch as the loss function.

We have run a hyperparameter search to find the best hyperparameters for the simple autoencoder
and the action-space autoencoder based on the loss $\L_{\text{Rec}}$ on the validation set.
To optimize the hyperparameters for PITA,
we have run a hyperparameter search
using the loss $\L$ as defined in \cref{eq:total_loss} on the validation set.
This hyperparameter search produced different models on the Pareto frontier
where the prediction performance qualitatively tended to go
either in the direction of better reconstruction or better alignment with the physical model.
We refer to these models as $\pita_{\text{Rec}}$ and $\pita_{\text{Phy}}$, respectively.
For the final training we fixed $m=5$ used the hyperparameters that produced the best results
in the hyperparameter search, which are provided in \cref{tab:hyperparameters}.
\begin{table}[t]
\centering
\begin{tabular}{lcccc}
\toprule
{Model} & \multicolumn{3}{c}{{Hyperparameters}} \\
& $\lambda_1$ $[10^{-4}]$ & $\lambda_2$ $[10^{-2}]$  &$\gamma$\\
\midrule
$\pita_{\text{Rec}}$ (ours) & 1.976 & 1.028 & 0.595 \\
$\pita_{\text{Phy}}$ (ours) & 1.030 & 3.012 & 0.032 \\
\bottomrule
\end{tabular}
\caption{Comparison of hyperparameters for the physics-informed models.}
\label{tab:hyperparameters}
\end{table}

\subsection{Comparison to Baseline Models}

To evaluate our approach we consider two orthogonal aspects:
\begin{enumerate}
	\item How well does the model reconstruct the input trajectory?
	\item How physically plausible is the reconstructed trajectory?
\end{enumerate}

\textbf{Reconstruction error:}

First and foremost, we compare how well our model is able to reconstruct the input trajectory
in comparison to the baseline models.
To this end, we measure the reconstruction in terms of the root mean squared error (RMSE)
between predicted and ground truth positions on the validation dataset.

\begin{figure}[t]
	\centering
	\includesvg[width=\linewidth]{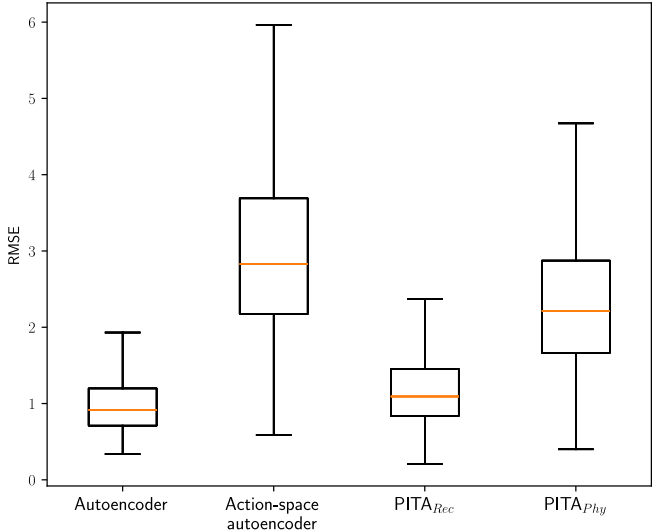}
	\caption{Reconstruction root mean squared error on the validation dataset.}
	\label{fig:reconstruction_error}
\end{figure}

As \cref{fig:reconstruction_error} illustrates, the simple autoencoder has the lowest reconstruction error,
closely followed by
$\pita_{\text{Rec}}$, which placed a higher emphasis on the reconstruction loss term.
Our $\pita_{\text{Phy}}$ model has a higher reconstruction error
because it focuses more on the physical loss term.
We can also observe, that the action-space autoencoder has the highest reconstruction error.

\textbf{Physical plausibility:}

To measure the adherence to the physical dynamics model, we experimented with different metrics.
For the first one, we compute the KBM state and input trajectory solely from the positions
predicted by the models using central finite differences.
That is, we approximately compute the input trajectory that is necessary to follow the predicted trajectory
with a KBM.
The resulting values for absolute acceleration and curvature are an indicator
for how physically plausible the predicted trajectory is.
Excessive use of control inputs compared to the ground truth indicates
that the predicted trajectory is not physically plausible.

\begin{figure}[t]
    \centering
    \begin{subfigure}{\linewidth}
        \centering
        \includesvg[width=\linewidth]{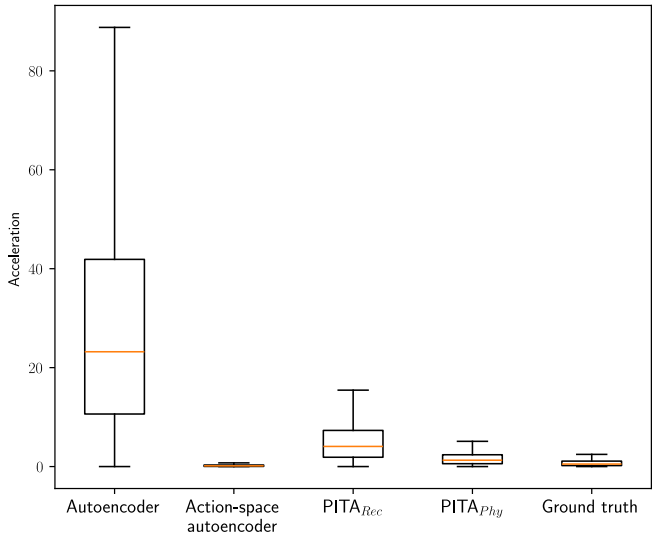}
        \caption{Absolute acceleration}
        \label{fig:model_acceleration}
    \end{subfigure}
    \hfill
    \begin{subfigure}{\linewidth}
        \centering
        \includesvg[width=\linewidth]{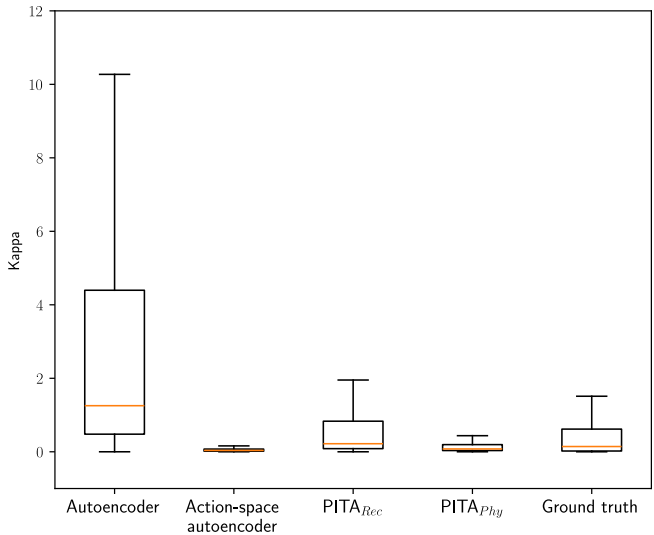}
        \caption{Absolute curvature}
        \label{fig:model_curvature}
    \end{subfigure}
    \caption{Boxplot of the model inputs necessary to follow the trajectories predicted by the different models wtih a KBM.}
    \label{fig:model_states}
\end{figure}

\Cref{fig:model_states} shows that the simple autoencoder model requires unrealistically high
model inputs to follow the predicted trajectory, indicating the poor
physical properties of its predictions.
On the contrary, the action-space autoencoder does not make use of excessive control inputs.
Both our PITA models are in between those two extremes,
where the model that put more weight on the physical loss term requires less control inputs.
While our models require more control inputs compared to the action-space autoencoder,
they need drastically less control inputs than the simple autoencoder.
We also show the computed model inputs for the ground truth trajectory as a reference.

As a second metric, we evaluate the smoothness of the predicted trajectory.
Predicted trajectories that are physically implausible because of noise or jitter
are expected to be less smooth than trajectories that do adhere to a physical motion model.
To measure the smoothness, we first compute a smoothed reference path
for every predicted trajectory using an Unscented Kalman Filter (UKF)
\cite{menegazSystematizationUnscentedKalman2015}
based on the KBM dynamics.
We then compute the average distance of the predicted positions to the smoothed reference path
as a measure for how smooth the predicted trajectory is.

\begin{figure}[t]
	\centering
	\includesvg[width=\linewidth]{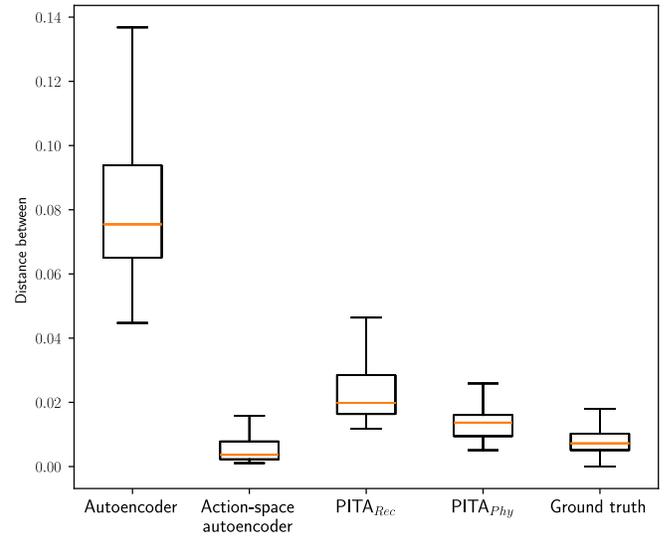}
	\caption{Boxplot of the distance of the predicted positions to the corresponding smoothed reference path.}
	\label{fig:smoothness}
\end{figure}

The results are displayed as a boxplot in \cref{fig:smoothness}.
We observe that the autoencoder model has the highest distance to the smoothed reference path,
indicating that the predicted trajectory is less smooth.
As expected, the action-space autoencoder has a very low distance to the smoothed reference path,
since it satisfies the KBM by design.
The two PITA models are again in between the two baseline models,
with the $\pita_{\text{Phy}}$ model having a lower distance to the smoothed reference path.
This is again expected, due to the higher weight placed on the physical loss term.
The values computed for the ground truth trajectory are between the
action-space autoencoder and $\pita_{\text{Phy}}$.

\section{Conclusion}

One problem with high fidelity trajectory autoencoders is that
due to the inherent noise they produce physically implausible trajectories.
To resolve this issue, we propose a novel approach to encode
and reconstruct vehicle trajectories using a combination of autoencoders
and physics-informed machine learning.
Our approach augments the reconstruction loss the a physics-based loss
that enforces the reconstructed trajectories to adhere to a model of the system dynamics.
We evaluate our approach on a real-world dataset and compare it
to a standard autoencoder and an action-space autoencoder.
The standard autoencoder produces the lowest reconstruction error
but also the noisiest trajectories.
On the contrary, the action-space autoencoder baseline achieves a smooth trajectory
but at the cost of a higher reconstruction error.
Our evaluation proves empirically that our approach can balance the two objectives and produce smooth and physically plausible trajectories.

To the best of our knowledge, this is the first work
to incorporate physics-informed machine learning principles
into trajectory encoding and prediction.
These concepts allow to learn an encoder that captures not only the future positions,
but also the dynamic properties on the system.
Based on the learned physics-informed encoder, we can train generative models
that produce physically plausible and realistic trajectories
for data augmentation.

In contrast to common prediction approaches,
our framework predicts output trajectory points at a high temporal resolution of
\SI{25}{\hertz}.
On the one hand this is necessary to augment data with high-fidelity.
On the other hand, this is prone to result in nonsmooth predictions due to the inherent prediction uncertainty.

Based on this work, our future research will focus on incorporating
additional context information on the traffic scene into the model.
This allows to encode the whole traffic scene instead of only
a single vehicle's trajectory.

{\small
\bibliographystyle{IEEEtran}
\bibliography{IEEEabrv,references_autoexport}
}

\newpage

\end{document}